\documentclass{article}
\usepackage{ijcai24}

\usepackage{times}
\usepackage{soul}
\usepackage{url}
\usepackage[hidelinks]{hyperref}
\usepackage[utf8]{inputenc}
\usepackage[small]{caption}
\usepackage{graphicx}
\usepackage{amsmath}
\usepackage{amsthm}
\usepackage{booktabs}
\usepackage{algorithm}
\usepackage{algorithmic}
\usepackage[switch]{lineno}
\usepackage{listings}

\usepackage{xspace}
\usepackage{acronym}
\usepackage{natbib}
\usepackage[capitalise]{cleveref}
\usepackage{amsmath}
\usepackage{amsfonts}
\usepackage{multirow}
\usepackage{makecell}
\usepackage[dvipsnames]{xcolor}

\acrodef{llm}[LLM]{Large Language Model}
\acrodef{3dvg}[3D-VG]{3D Visual Grounding}

\makeatletter
\DeclareRobustCommand\onedot{\futurelet\@let@token\@onedot}
\def\@onedot{\ifx\@let@token.\else.\null\fi\xspace}
\def\eg{\emph{e.g}\onedot} 

\def\ie{\emph{i.e}\onedot}

\def\etc{\emph{etc}\onedot}

\makeatother

\newcommand{\model}{\text{R2G}\xspace}
\newcommand{\supmat}{\textit{Sup. Mat.}\xspace}

\title{R2G: Reasoning to Ground in 3D Scenes}

\author{
    Yixuan Li, Zan Wang, Wei Liang
}

\author{
Yixuan Li$^1$
\and
Zan Wang$^1$\and
Wei Liang$^{1,2}$
\affiliations
$^1$Beijing Institute of Technology\\
$^2$Yangtze Delta Region Academy of Beijing Institute of Technology\\
}

\begin{document}
\maketitle

\begin{abstract}
We propose \textbf{R}easoning to \textbf{G}round (\model), a neural symbolic model that grounds the target objects within 3D scenes in a reasoning manner.
In contrast to prior works, \model \textbf{explicitly} models the 3D scene with a semantic concept-based scene graph; \textbf{recurrently} simulates the attention transferring across object entities; thus makes the process of grounding the target objects with the highest probability \textbf{interpretable}.
Specifically, we respectively embed multiple object properties within the graph nodes and spatial relations among entities within the edges, utilizing a predefined semantic vocabulary.
To guide attention transferring, we employ learning or prompting-based methods to analyze the referential utterance and convert it into reasoning instructions within the same semantic space.
In each reasoning round, \model either (1) merges current attention distribution with the similarity between the instruction and embedded entity properties or (2) shifts the attention across the scene graph based on the similarity between the instruction and embedded spatial relations.
The experiments on Sr3D/Nr3D benchmarks show that R2G achieves a comparable result with the prior works while maintaining improved interpretability, breaking a new path for 3D language grounding. Our project page is \url{https://sites.google.com/view/reasoning-to-ground}.
\end{abstract}

\section{Introduction} \label{sec:introduction}
Envision a scenario where we provide robots with several verbal instructions before leaving our residence, only to return and find that all the tasks have been completed. For robots to adhere to these instructions, they must possess the capability to comprehend language, perceive the 3D environment, and ground the objects mentioned in the language --- a sophisticated process known as \ac{3dvg} \citep{achlioptas2020referit_3d, chen2020scanrefer}. The preliminary \ac{3dvg} is crucial to empower machines to execute actions by locating the interacting furniture based on human instructions within intricate 3D scenes. Unlike humans, machines encounter significant challenges in comprehending the language and 3D scenes, let alone the difficulty of reasoning about the target object based on them.

Given a 3D scene and referential utterance, such as ``The bag on the couch,'' humans employ a form of attention-transferring to reason about the localization of the intended \textit{bag}.
They interpret the language description by decomposing it into several key clues: the \textbf{anchor} is the \textit{couch}, the \textbf{target} object is \textit{bag}, and the \textbf{spatial relation} between the anchor and the target is \textit{on}.
Guided by these clues, they first direct their attention to the \textit{couch} in the scene and then shift focus to the target \textit{bag}, recognizing its position as being \textit{on} the \textit{couch}.

\begin{figure}[t!]
    \centering
    \includegraphics[width=\linewidth]{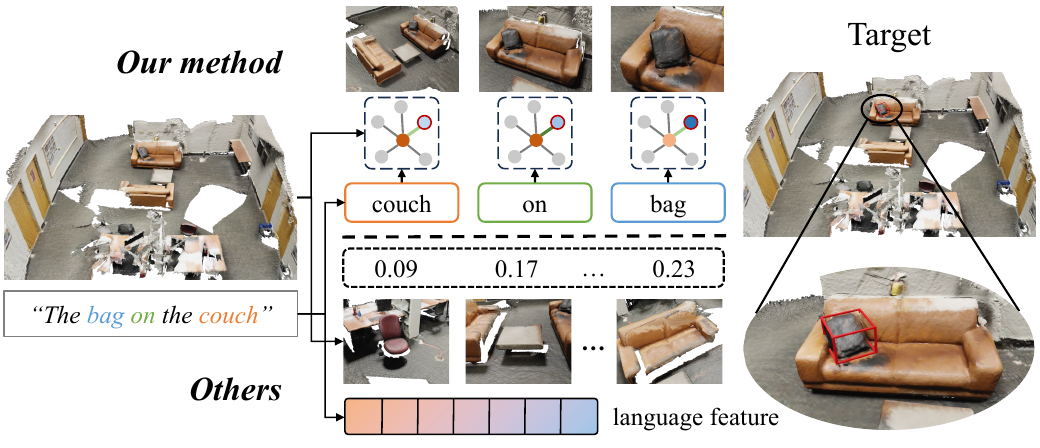}
    \caption{\textbf{Comparison between \model and previous models.} The prior works (bottom) focus on matching the utterance feature with the object proposal's features to select the target object with the highest probability in an end-to-end manner. In contrast, \model (top) grounds the target object step by step via human-like attention transferring across the scene graph, using the parsed language description as guidance.}
    \label{fig:teaser}
    \vspace{-12pt}
\end{figure}


In contrast to the explicit reasoning method employed by humans, recent studies on \ac{3dvg} have predominantly revolved around a detect-and-match methodology, as indicated at the bottom in \cref{fig:teaser}. These works fuse the features of language and object proposals using graph-based \citep{achlioptas2020referit_3d,huang2021text,yuan2021instancerefer} or transformer-based approaches \citep{yang2021sat,zhao2021_3DVG_Transformer,cai20223djcg,chen2023learning}.
Afterward, they feed the updated features into a classifier head to predict the probability of being referred to for each object. However, such an implicit paradigm that learns the alignment between 3D scene context and language descriptions leads to an uninterpretable grounding process, thus hindering generalization to novel scenes.
In a different approach, NS3D \citep{hsu2023ns3d} learns to filter out target object that satisfies the relation with the anchors, according to the structural programs translated by \ac{llm}.
Despite its advancements, NS3D faces two significant limitations: (1) It relies on MLPs to extract implicit representations for objects and relations and to check relations between objects, leading to uninterpretable reasoning and potentially inaccurate outcomes. (2) The absence of object attribute modeling poses a challenge in addressing attribute-related referential descriptions, \eg, "Find the white round table."

Drawing inspiration from human intelligence, we propose \model, a neural symbolic model, to ground the objects within 3D scenes using a more interpretable reasoning process, as shown in \cref{fig:teaser}. Specifically, we first build a semantic concept-based graph to represent the 3D scene through a predefined concept vocabulary. In this graph, individual nodes correspond to object entities, and edges represent spatial relationships between two entities. Subsequently, for every object proposal within the 3D scene, we employ an object classifier to predict its category and determine its attributes using heuristic rules. We further compute accurate spatial relations among entities heuristically by comparing the positions of proposals. Afterward, we map these semantic concepts, \ie, object properties and relations, to the concept vocabulary and embed them into GloVe space \citep{pennington2014glove}.

Secondly, we utilize RNN or \acp{llm} to parse the referential utterances into informative clues, dubbed as \textit{instructions}. Under the guidance of these instructions, we recurrently reason about the intended target object through the attention transferring on the scene graph. (1) For the instructions involving object properties, we compute the similarity between this instruction and the corresponding property embedded in nodes and merge the similarity with the current attention distribution. (2) For the instructions concerning spatial relations, we transfer the attention from the source to the target based on the similarity between the instruction and all relation concepts embedded in edges. After several reasoning rounds, \model finally grounds the target object with the highest attention score.

We evaluate our model on Sr3D/Nr3D \citep{achlioptas2020referit_3d} and NS3D \citep{hsu2023ns3d} benchmarks. The quantitative and ablative experiments show that \model achieves a comparable performance with the baseline methods, thus affirming the effectiveness of our proposed representation and reasoning paradigm. Furthermore, qualitative analysis of the reasoning process highlights our model's enhanced interpretability and generalization ability.

The primary contributions of this work are three-fold: (1) we propose the first neural symbolic model for \ac{3dvg}, capable of processing both relation-oriented and attributed-related referential utterances; (2) we heuristically compute the spatial relation and object attributes from the 3D point cloud and represent them with explicit semantic concepts; (3) the extensive experiments on Sr3D/Nr3D benchmark show that \model can achieve a comparable result with the prior works while maintaining improved interpretability and generalization ability.

\section{Related Work} \label{sec:related work}

\paragraph{Visual Grounding}
The goal of visual grounding is to localize the target object according to referential utterances in either a provided image, known as 2D grounding \citep{hu2016natural,sadhu2019zero,yang2020improving,liao2020real}, or within a 3D scene, \ie, 3D grounding \citep{achlioptas2020referit_3d,chen2020scanrefer}.
Previous methods in 2D grounding solve this task using two mainstream frameworks: one-stage methods and two-stage methods. One-stage methods \citep{sadhu2019zero,yang2020improving,liao2020real} directly regress the bounding box for the target object, while two-stage methods \citep{deng2018visual,yang2019dynamic,chen2021ref} first generate the candidate object region proposals and then match the object's feature with linguistic feature to identify the target object. 
Likewise, \citep{luo20223d, chen2023learning} explore single-stage pipelines for \ac{3dvg}.
Two-stage methods either use the graph neural network to extract spatial features among proposals \citep{achlioptas2020referit_3d,huang2021text,feng2021free,yuan2021instancerefer} or adopt Transformer architecture to integrate object features with linguistic features \citep{zhao2021_3DVG_Transformer,he2021transrefer3d,huang2022multi,luo20223d,cai20223djcg,chen2021d3net,chen2022vil3dref,bakr2022look}, following the stage of generating object proposals by 3D detection backbones \citep{qi2017pointnet++}. \citet{yang2021sat,huang2022multi} further leverage 2D information to amplify the representation of proposals, thereby bolstering the grounding process.
However, all the above works ground the target object end-to-end, thus hindering the model's interpretability and generalization ability.
\citet{hsu2023ns3d} introduce the first neural-symbolic model to address \ac{3dvg} in a reasoning manner, but limited to implicit representation and uninterpretable relation checking, and neglecting attribute-related descriptions. Our proposed method grounds the targets through explicit reasoning on a semantics-rich graph embedded with object properties and spatial relationships, supporting both spatial-relation-oriented and attribute-related descriptions.

\paragraph{Visual Reasoning}
In recent years, following some efforts in 2D Visual Question Answering (VQA) \citep{anderson2018bottom,yu2018beyond,hudson2019gqa,cao2019interpretable}, \citet{ma2022sqa3d,azuma2022scanqa,ye20223d} introduce analogous visual reasoning tasks into the domain of 3D. Despite the significant advancements, most works directly deduce the answer from a collection of implicit features extracted from visual and linguistic inputs, thereby bypassing the inherent logical reasoning within the problem.
\citet{hudson2019learning} introduces a novel neural symbolic model to solve the 2D VQA task by instruction-guided reasoning within a neural state machine. NS3D \citep{hsu2023ns3d} integrates the power of large language models and modular neural networks to reason about the spatial relations between objects. In this work, we introduce \model to represent the 3D scene and referential utterances with explicit semantic concepts and localize the target object through interpretable simulation of attention transferring across the scene graph.

\section{Method} \label{sec:method}

\begin{figure*}[t!]
    \centering
    \includegraphics[width=\linewidth]{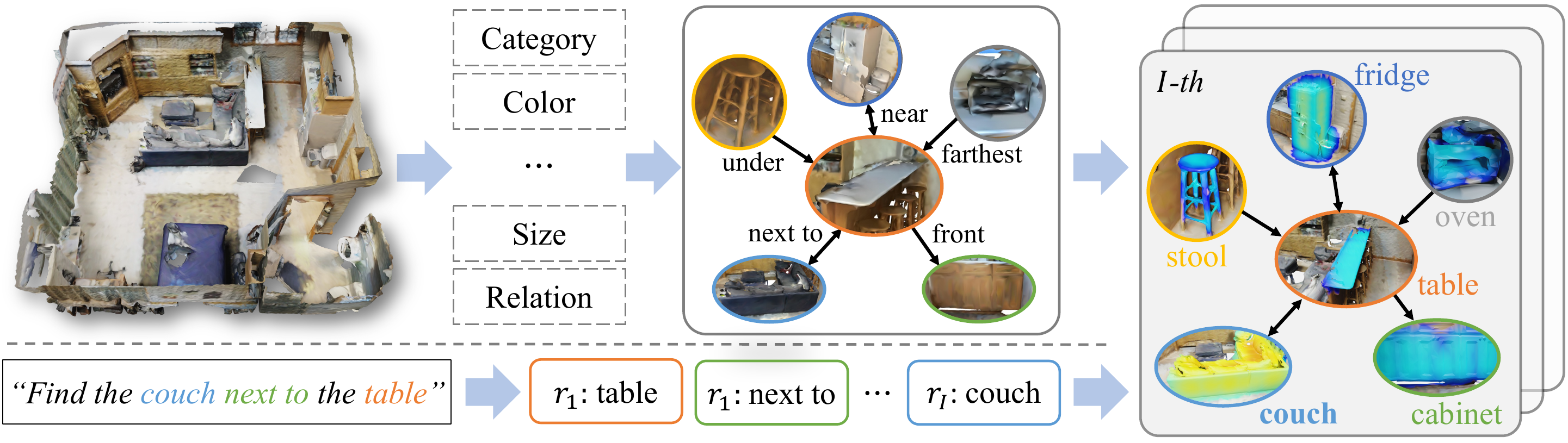}
    \vspace{-16pt}
    \caption{\textbf{Overview of \model.} \model represents the 3D scene with a semantic concept-based scene graph and parses the referential utterance into instructions to guide the attention transferring across the scene graph in a reasoning manner. After several reasoning rounds, we localize the target object with the highest attention score.}
    \label{fig:pipeline}
    \vspace{-12pt}
\end{figure*}

Given a referential language description, the objective of \ac{3dvg} is to localize the target object in the 3D scene. In this work, we propose a novel neural-symbolic method that interpretably grounds the target object in a reasoning manner, as depicted in \cref{fig:pipeline}. In this section, we will introduce the semantic concept-based representation of the 3D scene and the language description, as well as the rationale behind the grounding process.

\subsection{Semantic Representation} \label{sec:representation}

Unlike the prior works using the fused latent features, \model represents the 3D scene and language description with shared explicit semantic concepts.
We first establish a semantic concept vocabulary $C=C_O \cup C_A \cup C_R$, where $C_O$ contains the semantic concepts related to object categories, such as ``table'' and ``chair'', $C_A$ contains the attribute-related concepts like ``red'' and ``round'', and $C_R$ contains the semantic concepts about the spatial relationships between objects, such as ``near'' and ``bellow''. We embed these concepts into GloVe space \cite{pennington2014glove}, where each concept is initialized with a $d$-dimensional embedding. By translating both the visual and linguistic information into such explicit semantic concepts, we are able to reason about the relevance between the two modalities (\ie, computing the similarity of embedded concepts), facilitating high-level abstract reasoning and improving the model's interpretability.

\subsection{Scene Graph Construction} \label{sec:scene graph}

We represent the given 3D scene using a semantic concept-based scene graph, denoted as $\mathcal{G} = \{\mathcal{S}, \mathcal{E}\}$. Within the scene graph, each node encapsulates the state of a 3D object encompassing its category and attribute, and each edge represents a spatial relation existing between two objects, as depicted in \cref{fig:pipeline}.
Such a scene graph comprises two fundamental components: a node state set denoted as $\mathcal{S}$ and an edge set denoted as $\mathcal{E}$.
Next, we will introduce the methodology employed for embedding object properties into the node state $s \in \mathcal{S}$ and spatial relation into the edge $e \in \mathcal{E}$.

\paragraph{Object State} In the scene graph, each object node state $s$ comprises $L+1$ embedding, denoted as $\{s^j\}_{j=0}^L$. Here, $s^0$ represents the object category embedding, while $s^1, \cdots, s^L$ correspond to the embeddings of $L$ object attributes.

For a given 3D scene point cloud, we follow the previous setting that assumes to be given a list of $N$ object proposals $\{O_i\}_{i=1}^{N}$. These proposals are acquired either through ground truth annotation \citep{achlioptas2020referit_3d} or 3D instance segmentation \citep{chen2020scanrefer}.
Subsequently, we utilize pre-trained PointNet++ \citep{qi2017pointnet++} for object category regression, yielding a probability distribution $P(\cdot)$ over pre-defined object categories.
We compute the category embedding of each object proposal as the weighted summation of semantic concepts derived from the predicted $P(\cdot)$, formulated as
\begin{equation}
    s^{0}=\sum_{c_k \in C_O}P(k) c_k.
    \label{eq:cate_embedding}
\end{equation}
$c_k$ is the concept embedding most relevant to category $k$.

Considering that the referential utterances may encompass attribute-related descriptions, we also incorporate the object attributes into the node state $s$. We model $L=9$ attribute types for each object, including color, shape, size, \etc. The respective attribute embeddings $\{s^j\}_{j=1}^L$ are derived from the attribute concept set $C_A = \cup_{i=i}^{L} C_{i}$.
Specifically, for object color (\eg, red, white, and others), we employ a heuristic approach to choose the most relevant color concept embedding, informed by the proposal's point RGB values.
In the case of object shape (\eg, round and square), we adopt an MLP to predict object shape types from PointNet++ features.
This process yields a shape embedding that is computed as a weighted summation of shape concepts embedding, similar to the formulation presented in \cref{eq:cate_embedding}.
Meanwhile, object attributes like size (e.g., biggest and smallest) necessitate inter-object category comparisons. To address this, we compute embeddings for these attributes based on object category distribution and the size of bounding boxes. The approach is analogous to the computation of spatial relation embedding for ``farthest'' and ``closest'', which we will explain in detail in upcoming sections.
We direct readers to \supmat for more details about the embedding computation of all object properties.

\paragraph{Spatial Relation} We embed each directed edge $e$ that links two object nodes using the spatial relation concepts in $C_R$. It's noteworthy that each edge can encompass multiple relationships $\{e^{j}\}_{j=1}^{T}$. For instance, a coffee table could simultaneously be positioned closest to and in front of a sofa.
Here, $e^{j}$ is the most relevant concept embedding related to $j$-th relation type. In this work, we model $T=10$ distinct spatial relation types, most of which can be deduced from the positional coordinates and orientations of object proposals through heuristic rules. In cases where such relations exist, we set the probability of these relations as $R(\cdot) = 1$; otherwise, $R(\cdot) = 0$.

Since ``farthest'' and ``closest'' are category-dependent relationships necessitating comparisons among objects within the identical category, and considering that the object category possesses a probability distribution, it becomes crucial to determine the relation probability $R(\cdot)$.
In particular, for each object pair $(z, x)$, we compute the probability that ``$x$ is farthest (or closest) to $z$ when $x$ belongs to category $k$'' as follows:
\begin{equation}
    \small
    r_k(z, x) = P_x(k)\sum_{Y \in \mathcal{Y}} \prod_{y \in Y} \prod_{\bar{y} \in \bar{Y}}  \mathbb{I}(z, x, y) P_y(k) (1 - P_{\bar{y}}(k)),
\end{equation}
where $\mathcal{N} = \{1, \cdots, N\} / \{z, x\}$ is the object set excluding $z$ and $x$, $\mathcal{Y}$ is the power set of $\mathcal{N}$, $\bar{Y} = \mathcal{N} / Y$, and $\mathbb{I}(\cdot)$ is a indicator function, indicating ``$x$ is farther (or closer) to $z$ than $y$'', formulated as
\begin{equation*}
    \small
    \mathbb{I}(z, x, y) =
        \begin{cases}
            1 & x\ \text{is farther (or closer) to}\ z\ \text{than}\ y \\
            0 & else.
        \end{cases}
    \label{eq:indicator}
\end{equation*}
The probability $R(z, x)$ for ``$x$ is farthest (closest) to $z$'' is the summation of $r_k(z, x)$ across all categories.
To save computation, we only consider the top $K$ categories with the highest probability for each object.
At length, the edge embedding is computed as $e = \sum_{j=1}^T R^j e^j$. We set $K=2$ in our main experiments.

We recommend referring to the \supmat for more details about the construction of the scene graph.

\subsection{Instruction Generation} \label{sec:instruction}


Given a referential utterance, such as ``find the black couch next to the table,'' humans naturally dissect this description into distinct semantic directives,
guiding us first to localize the ``table'' (\textit{anchor}), followed by transferring our focus from the ``table'' to the black ``couch'' (\textit{target}), facilitated by the intervening relation ``next to'' (\textit{relation}).
Inspired by this cognitive process, we introduce two methodologies to convert the provided language description into a series of instructions denoted as $\{r_i\}_{i=1}^{I}$, \ie, a learning-based approach and an LLM-based approach.

\paragraph{Learning-based} Following \citet{hudson2019learning}, we embed all utterance words with GloVe \citep{pennington2014glove}; replace each word with the most relevant concept embedding defined in $C$ or keep it if there is no matching concept. For each word embedding $w_i$, we compute a similarity distribution $P_i^w$ across the vocabulary $\widehat{C}$ as follows:
\begin{equation}
    P_i^w = \text{softmax}(w_i^T \mathbf{W}_w \widehat{C}),
\end{equation}
where $\mathbf{W}_w$ is a learnable matrix and $\widehat{C}$ is the union of $C$ and $\{c'\}$. $c'$ is a learnable embedding representing no-content words.
We then replace each word embedding with $v_i = \sum_{c \in \widehat{C}} P_i^w(c) c$.
By forwarding the word embeddings $V^{\Omega\times d}=\{v_i\}^{\Omega}_{i=1}$ into an RNN-based encoder-decoder sequentially, we roll out the decoder for $I$ times to generate $I$ hidden states $\{h_i\}_{i=1}^{I}$ and transform them into instructions as follows:
\begin{equation}
    r_i = \text{softmax}( h_iV^T )V.
\end{equation}

\paragraph{LLM-based} By leveraging the power of \acp{llm} like ChatGPT \citep{chatgpt}, we prompt these models to parse the referential utterance into key clues, including properties of \textit{anchor} and \textit{target}, along with the spatial \textit{relation}. We represent the parsed results with the most relevant concepts in $C$ and regard the corresponding embeddings as instructions. Furthermore, we employ a ``zero'' embedding to pad the corresponding instructions when properties of the \textit{target} and \textit{anchor}, or the spatial relationship, cannot be extracted from the given referential utterance.

Moreover, we have different choices for the instruction number $I$ for different evaluation benchmarks.
Since the benchmark Sr3D \citep{achlioptas2020referit_3d} only involves spatially-oriented descriptions and neglects objects' attributes, we set $I = 3$ for guiding the model to reason about the \textit{target} category, the spatial \textit{relation}, and the \textit{anchor} category with these three instructions, respectively.
To handle attribute-related descriptions, we generate $I = 2 L + 3$ instructions, which we explicitly align with $L + 1$ properties of the \textit{target}, $L + 1$ properties of the \textit{anchor}, and $1$ spatial \textit{relation} connecting the target and anchor.
Please refer to \supmat for more details about the instruction generation.

\subsection{Reasoning} \label{sec:reasoning}

\begin{figure}[t!]
    \centering
    \includegraphics[width=0.75\linewidth]{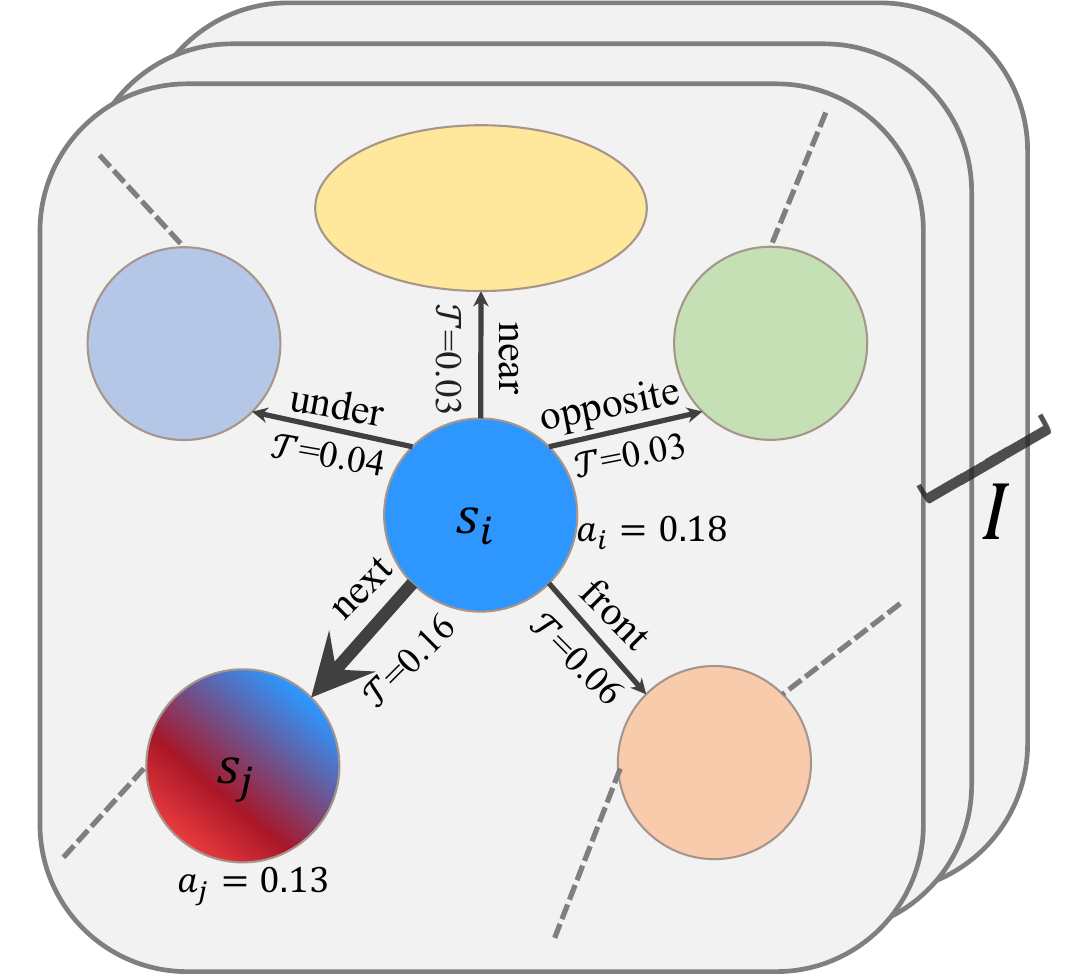}
    \vspace{-4pt}
    \caption{\textbf{Attention transferring.} \model transfers the attention from the source node to the target node along the directed edge, guided by the spatial-relation-related instruction.}
    \label{fig:transfer}
    \vspace{-8pt}
\end{figure}

To localize the target object, we leverage the generated instructions $\{r_i\}_{i=1}^I$ to guide the attention to transfer across the scene graph $\mathcal{G}$ in a reasoning manner. The process begins with a uniform attention distribution over the nodes, denoted as $\{a_0^s = \frac{1}{N}\}_{s \in \mathcal{S}}$. Subsequently, over a series of $I$ reasoning rounds,  we re-distribute the attention across the scene graph from $a_{i-1}$ to $a_{i}$ with the guidance of $r_{i}$.

For the first $L + 1$ instructions, we assume that these instructions are relevant to the \textit{anchor}'s properties, \ie, category, color, shape, \etc, respectively. We compute an attention distribution $\{b_i^s\}_{s \in \mathcal{S}}$ through the similarity between each instruction and the corresponding property concepts embedded in the graph node, formulated as,
\begin{equation}
    b_{i}^s = \text{softmax}_{s \in \mathcal{S}}(\mathbf{W}_s \cdot \sigma (r_i \circ \mathbf{W}^j s^j)),
    \label{eq:attn_instr}
\end{equation}
where $\mathbf{W}_s$ and $\mathbf{W}^j$ are learnable parameters, $\sigma$ is a non-linear projection, $\circ$ is Hadamard product, and $j = i - 1 = 0, \cdots, L$.
In these reasoning rounds, we merge the one-step predicted attention distribution $\{b_i^s\}_{s \in \mathcal{S}}$ with pre-step distribution $\{a_{i-1}\}_{s \in \mathcal{S}}$ to generate $\{a_{i}\}_{s \in \mathcal{S}}$, \ie,
\begin{equation}
    a_{i}^s = \text{softmax}_{s \in \mathcal{S}} (b_i + a_{i - 1}).
    \label{eq:attn_merge}
\end{equation}

We assume that the $(L + 2)$-th instruction is relevant to the spatial relation between the \textit{target} and \textit{anchor}. In this reasoning round, we transfer the attention from the \textit{anchor} to the \textit{target} via the similarity between the instruction $r_{L+2}$ and the relationship concepts embedded in $\{e^{i}\}_{i=1}^{T}$, as illustrated in \cref{fig:transfer}. The attention distribution in this reasoning round is updated as
\begin{equation}
    a_{L + 2}^s = \text{softmax}_{s \in \mathcal{S}}(\mathbf{W}_r \cdot \sum_{(s', s) \in \mathcal{E}} p_i(s') \cdot \mathcal{T}((s', s))),
\end{equation}
where $\mathcal{T}((s', s))$ represents the amount of attention transferred from entity node $s'$ to $s$ through edge $(s', s)$, with its embedding denoted as $e'$.
This is computed as follows:
\begin{equation}
    \mathcal{T}(s', s) = \sigma(r_{L+2} \circ \mathbf{W}_e e'),
    \label{eq:attn_transfer}
\end{equation}
where $\mathbf{W}_r$ and $\mathbf{W}_e$ are learnable parameters.

Similarly, we assume the last $L + 1$ instructions are relevant to the properties of \textit{target}. We use \cref{eq:attn_instr} and \cref{eq:attn_merge} to update the attention distribution in the last $L + 1$ reasoning rounds. At length, we localize the target object with the highest attention score.

\subsection{Training Loss} \label{sec:loss}

To train \model, we mainly employ a grounding loss $\mathcal{L}_{ref}$ formulated as a cross-entropy loss over $N$ objects. For the model that parses descriptions using the learning-based method, we introduce auxiliary loss functions to regress corresponding concepts from the parsed instructions, \eg, the target object categories. These auxiliary loss functions are formulated as a cross-entropy loss as well. Notably, the auxiliary loss functions can only be used in the benchmark involving the annotations for language descriptions.

To this end, we train our model on Sr3D \citep{achlioptas2020referit_3d} with the combination of $\mathcal{L}_{ref}$ and three auxiliary loss functions. The loss function is formulated as
\begin{equation}
    \label{eq:loss}
    \begin{split}
        L=\mathcal{L}_{ref} + \alpha_{1}\mathcal{L}_{t} + \alpha_{3}\mathcal{L}_{a} + \alpha_{2}\mathcal{L}_{r}.
    \end{split}
\end{equation}
We empirically set the hyper-parameters $\alpha_{1}=0.2$, $\alpha_{2}=0.2$, and $\alpha_{3}=0.2$. $\mathcal{L}_{t}$, $\mathcal{L}_{a}$, and $\mathcal{L}_{r}$ are designed for regressing the target, anchor object category, and the spatial relation type from corresponding instructions.

\subsection{Implementation Details}
The model is trained on an NVIDIA RTX 3090Ti GPU with PyTorch framework in an end-to-end way. We set the initial learning rate as 1e-4 and reduce it if there is no improvement in grounding accuracy after ten epochs. We use the Adam optimizer and train each model until convergence. 

\section{Experiment} \label{sec:experiment}

\begin{table}[ht!]
    \centering
    \caption{\textbf{Quantitative results on Sr3D \citep{achlioptas2020referit_3d}.}}
    \vspace{-6pt}
    \label{tab:result_sr3d}
    \resizebox{\linewidth}{!}{%
        \begin{tabular}{ccc}
        \toprule
        \multirow{2}{*}{Method} & \multicolumn{2}{c}{Sr3D} \\
         & Overall & View-dep. \\
        \hline
        ReferIt3D \citep{achlioptas2020referit_3d}          & $40.8\%$ & $39.2\%$ \\
        TGNN \citep{huang2021text}                          & $45.0\%$ & $45.8\%$ \\
        InstanceRefer \citep{yuan2021instancerefer}         & $48.0\%$ & $45.4\%$ \\
        SAT \citep{yang2021sat}                             & $57.9\%$ & $49.2\%$ \\
        3DVG-Transformer\citep{zhao2021_3DVG_Transformer}   & $51.4\%$ & $44.6\%$ \\
        TransRefer3D \citep{he2021transrefer3d}             & $57.4\%$ & $49.9\%$ \\
        MVT \citep{huang2022multi}                          & $64.5\%$ & $58.4\%$ \\
        3D-SPS \citep{luo20223d}                            & $62.6\%$ & $49.2\%$ \\
        BUTD-DETR \citep{jain2022bottom}                    & $67.0\%$ & $53.0\%$ \\
        \hline
        NS3D \citep{hsu2023ns3d}                            & $62.7\%$ & $62.0\%$ \\
        Ours (\model + RNN)                                 & $61.2\%$ & $74.8\%$ \\
        Ours (\model + \ac{llm})                             & $55.7\%$ & $68.7\%$ \\
        \bottomrule
        \end{tabular}%
     }%
\end{table}

In this section, we will demonstrate the effectiveness of our model in comparison to other baselines through a series of quantitative experiments.
Furthermore, the conducted ablation studies clarify how individual modules contribute to the overall performance.

\subsection{Experimental Setting}

\paragraph{Datasets}
We evaluate our method on 3D grounding benchmarks ReferIt3D \citep{achlioptas2020referit_3d} and NS3D \citep{hsu2023ns3d}.
ReferIt3D consists of two subsets, \ie, Nr3D and Sr3D, which are derived from 707 ScanNet indoor scenes \citep{dai2017scannet}. The former has 83k utterances generated by the ``\text{\textcolor{Cyan}{target-class}}''--``\text{\textcolor{ForestGreen}{spatial-relation}}''--``\text{\textcolor{YellowOrange}{anchor-class(es)}}'' template. The latter has 41k natural language descriptions collected from humans with an online reference game.
ReferIt3D further split the datasets into ``Easy'' and ``Hard'' sets according to the difficulty level of utterances and into ``View-dependent'' and ``View-independent'' sets based on whether the descriptions are dependent on the view direction or not.
NS3D dataset is a sub-dataset of Nr3D that consists of 3659 examples for training and 1041 for evaluation, in which all the referential utterances describe a spatial relationship between an anchor and a target object.

\paragraph{Evaluation Metric}
Following the settings in the previous works, we use grounding accuracy as the metric.

\subsection{Results}

\begin{figure*}[t!]
    \centering
    \includegraphics[width=\linewidth]{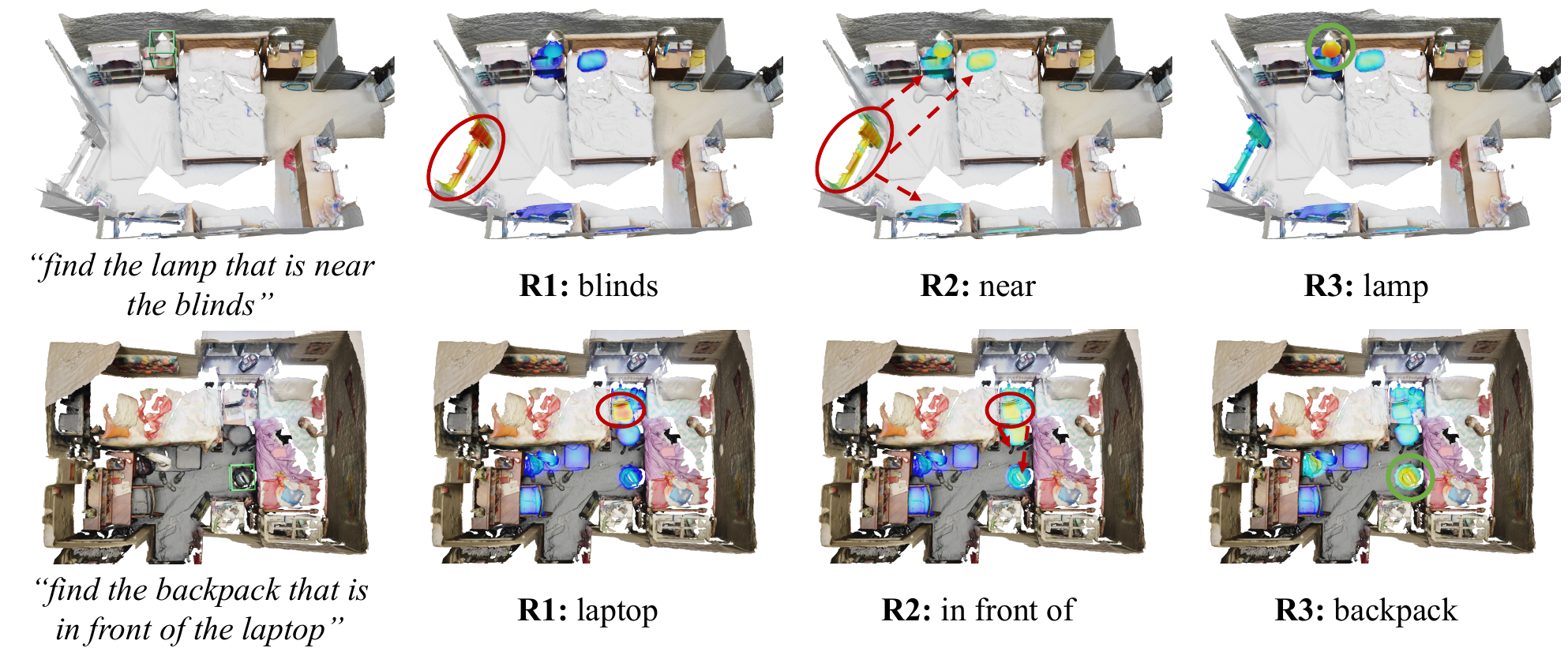}
    \vspace{-20pt}
    \caption{\textbf{Qualitative results.} We visualize two examples of the attention-transferring process on Sr3D in three reasoning rounds. \model gradually focuses more on the target object. We visualize the attention score of partial objects for better visualization.}
    \label{fig:qualitative}
    \vspace{-12pt}
\end{figure*}

\begin{figure}[t!]
    \centering
    \includegraphics[width=\linewidth]{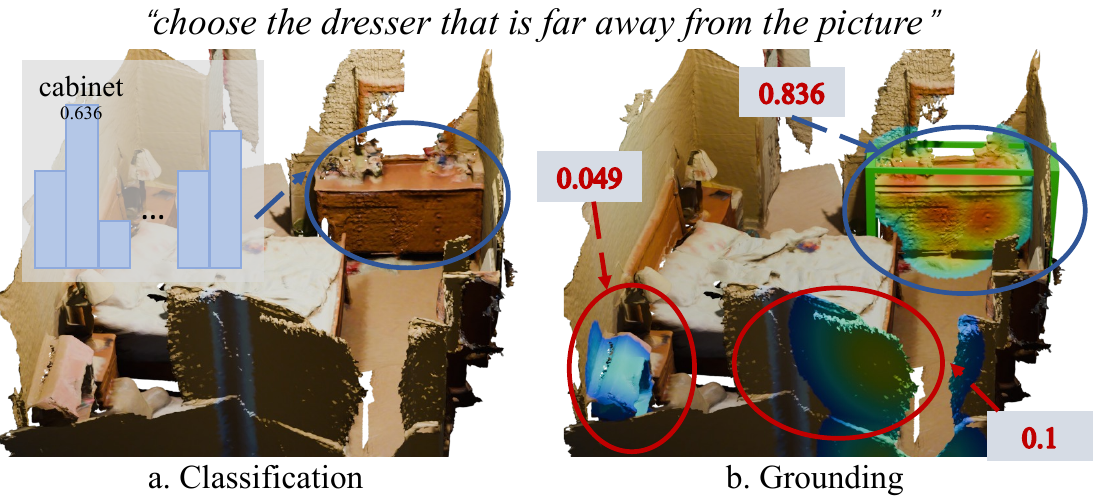}
    \vspace{-20pt}
    \caption{\textbf{Qualitative analysis of the end-to-end model.} The end-to-end model produces an erroneous classification (a) but successfully achieves a correct grounding result (b).}
    \label{fig:qualitative_analysis}
    \vspace{-12pt}
\end{figure}

\paragraph{Quantitative Results}
\cref{tab:result_sr3d} presents a comparative analysis of grounding accuracy between our model and other baselines in Sr3d and Nr3d datasets. The upper part is the result of conventional end-to-end approaches, and the lower part is the result of neural symbolic approaches.
Our model achieves an overall accuracy of $61.2\%$ in the Sr3D dataset, a result that is comparable with end-to-end approaches while maintaining a significant improvement in the model's interpretability.
Additionally, \model achieves the state-of-the-art accuracy of the ``View-dependent'' set, surpassing the prior state-of-the-art performance by $12.8\%$ grounding accuracy.
This outcome further demonstrates that \model can identify spatial relations within the scene graph and precisely localize the target, facilitated by a meticulously constructed scene graph.

We expand the scope of our model to process human linguistics using \ac{llm} within the Nr3D dataset, resulting in a grounding accuracy of approximately $25.8\%$, as shown in \cref{tab:result_nr3d}.
Owing to the absence of attributive modeling, NS3D \citep{hsu2023ns3d} achieves only a $7.3\%$ overall grounding accuracy, significantly trailing behind our approach. The results demonstrate that \model works better when coping with straightforward natural languages than NS3D.

We also perform a comparative experiment on the NS3D dataset, a subset of the Nr3D dataset, using the same experimental settings.
We directly apply our \ac{llm}-based model trained on Sr3D to achieve a result comparable to N3D \textbf{without} any fine-tuning. The results are shown in \cref{tab:result_nr3d}.

\paragraph{Qualitative Results} \cref{fig:qualitative} shows the grounding process of \model through two examples in Sr3D dataset. Across three reasoning rounds, \model initially focuses on the anchor objects, subsequently shifts its attention to other objects as guided by the spatial relationship, and enhances the target's attention score via the last instruction. \model finally identifies the target object with the highest attention score, thereby enhancing the interpretability of the grounding process.

In addition, we qualitatively analyze the grounding process of the previous end-to-end models, as exemplified in \cref{fig:qualitative_analysis}. The model initially classifies the target object \textit{dresser} as a \textit{cabinet}, yet ultimately yields accurate grounding results. We speculate that the model might learn correlations that do not imply causal relations, such as directly grounding the farthest object if its distance to the anchor is above a specified threshold. This phenomenon highlights the limited interpretability of such end-to-end models, thus leading to unreliable results.

\begin{figure*}
    \centering
    \includegraphics[width=0.98\linewidth]{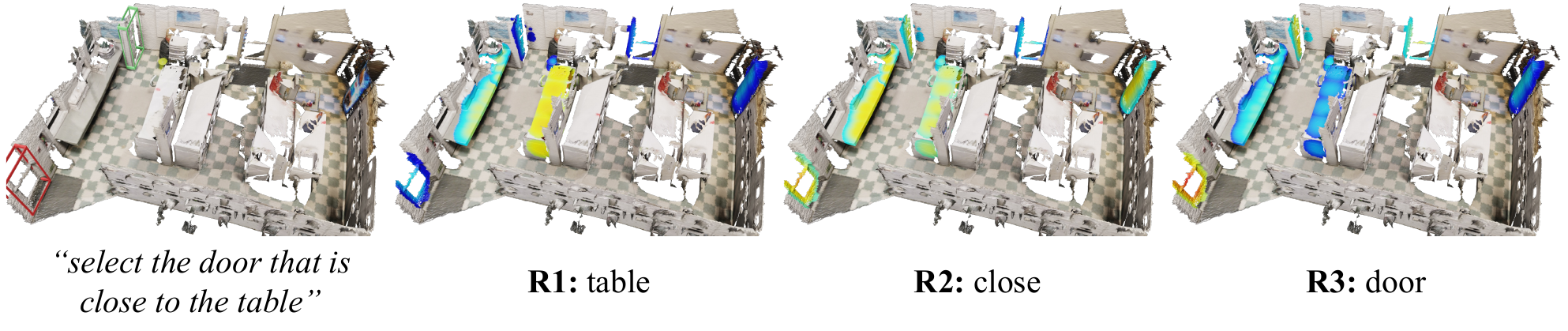}
    \vspace{-6pt}
    \caption{\textbf{Failure case.} \model fails to localize the target ``door'' with a green bounding box, but the ``door'' in a red bounding box due to the wrong localization of the anchor ``table'' and the inaccurate object classification results.}
    \label{fig:failure}
    \vspace{-12pt}
\end{figure*}

\paragraph{Failure Case}
We present a typical failure case in \cref{fig:failure}. In the first column, the green bounding box corresponds to the target we need to localize, while the red bounding box represents the outcome of our model. The rest of the figures show the attention-transferring process, in which the model initially erroneously localizes the object ``washing machines'' as a table with the anchor instruction ``table'', due to a wrong object classification outcome. Subsequently, the model transfers the attention from the incorrect anchor object, eventually resulting in a wrong grounding result with the target object erroneously classified as a ``wall''.

\begin{table}[t!]
    \centering
    \caption{\textbf{Quantitative results on Nr3D.}}
    \vspace{-6pt}
    \label{tab:result_nr3d}
    \resizebox{\linewidth}{!}{%
        \begin{tabular}{cccc}
        \toprule
        \multirow{2}{*}{Method} & \multicolumn{2}{c}{Nr3D} & NS3D \\
        & Overall & View-dep. & Overall \\
        \hline
        NS3D \citep{hsu2023ns3d} & $7.3\%$  & $8.5\%$  & $52.6\%$\\
        Ours (\model + RNN)      & $13.9\%$ & $13.9\%$ & $46.8\%$\\
        Ours (\model + \ac{llm}) & $25.8\%$ & $24.0\%$ & $50.8\%$\\
        \bottomrule
        \end{tabular}%
     }%
     \vspace{-6pt}
\end{table}


\subsection{Ablation Study} \label{sec:ablation}

\begin{figure}[t!]
    \centering
    \includegraphics[width=\linewidth]{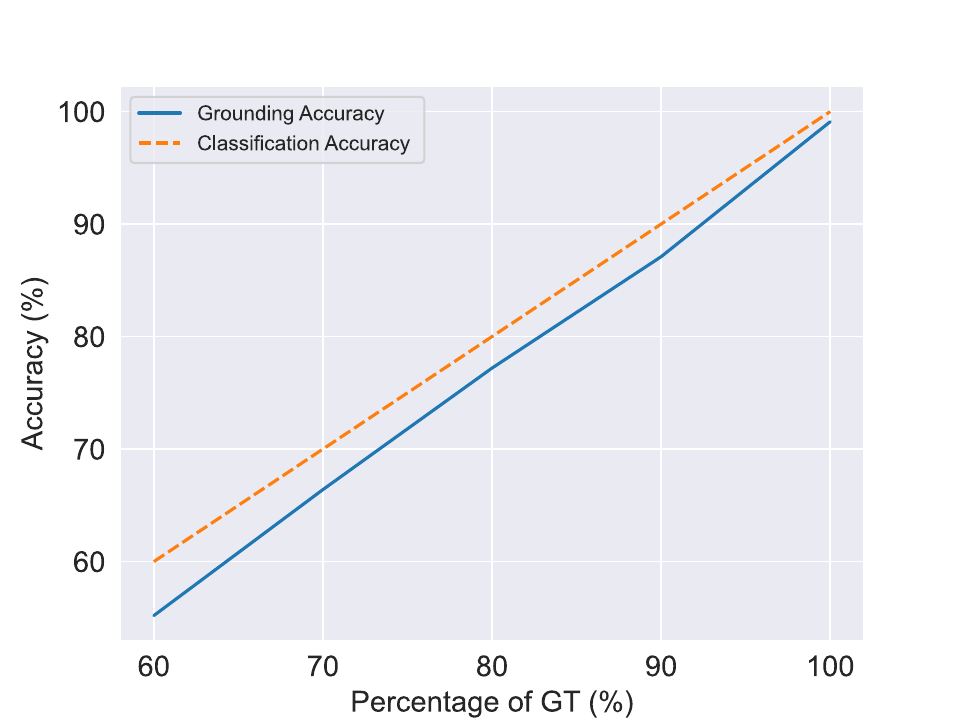}
    \vspace{-16pt}
    \caption{\textbf{Grounding performance varying with the proportion of the object category ground truth.}}
    \label{fig:acc-curve}
    \vspace{-4pt}
\end{figure}

\paragraph{Object Classification Accuracy}
We evaluate our model's performance on the Sr3D dataset while excluding the "between" relationship to examine the impact of object category classification accuracy on grounding performance. We first conduct experiments to train \model with and without using the ground truth object category. As depicted in \cref{tab:result_sr3d_gt}, our model trained with the ground truth object categories showcases a notable enhancement in grounding performance compared to its counterpart without such GT. Remarkably, our model trained with GT achieves a surprising grounding accuracy, \ie, $\mathbf{99.1}\%$, significantly outperforming the NS3D model. Moreover, we systematically vary the object classification accuracy to observe its influence on grounding performance. Specifically, we randomly assign categories to partial objects in each scene to adjust the overall accuracy of the object category, simulating an ascending performance trend in object classification.
As illustrated in \cref{fig:acc-curve}, the outcomes demonstrate a clear positive correlation: higher object classification accuracy results in heightened grounding accuracy. This observation accentuates that our model's current performance bottleneck primarily resides in the precision of object classification. By enhancing the capabilities of the classification head, our model's performance could improve further.

\begin{table}[t!]
    \centering
    \caption{\textbf{Grounding accuracy of the neural-symbolic method on Sr3D without ``between'' samples.} ``w/ GT'' and ``w/o GT''  indicate whether we use the ground truth object category.}
    \vspace{-6pt}
    \label{tab:result_sr3d_gt}
    \resizebox{0.82\linewidth}{!}{%
        \begin{tabular}{ccc}
        \toprule
        \multicolumn{2}{c}{Method} & Sr3D \\
        \hline
        \multirow{2}{*}{w/o GT} & NS3D \citep{hsu2023ns3d} & $61.9\%$ \\
                                & \model(ours)             & $\mathbf{62.2\%}$ \\
                                
        \hline
        \multirow{2}{*}{w/ GT}  & NS3D \citep{hsu2023ns3d} & $94.0\%$ \\
                                & \model(ours)             & $\mathbf{99.1\%}$ \\  
        \bottomrule
        \end{tabular}%
    }%
    \vspace{-6pt}
\end{table}

\paragraph{Top $K$ Object Category}
As illustrated in \cref{sec:scene graph}, we only consider the top $K$ object categories with the highest probability for computation efficiency instead of using the probability distribution over all categories.
To this end, we delve into how the selection of $K$ directly influences the grounding accuracy.
By evaluating our model with $K=1$ and $K=2$ on the Sr3D dataset, excluding the ``between'' relationship, we achieve $\mathbf{55.6\%}$ and $\mathbf{61.3\%}$ grounding accuracy, respectively.
The performance improvement from $K=1$ to $K=2$ indicates that the top $1$ model only leverages the object category with the highest probability to compute the ``farthest'' and ``closest'' relationship, potentially disregarding other possible categories. The top $2$ model leverages more object categories over a probability distribution, thus resulting in a more accurate probability for corresponding relationships.

\paragraph{Loss Function}
We further investigate the importance of the introduced auxiliary loss functions within the Sr3D benchmark. As shown in \cref{tab:loss_ablation}, our full model surpasses all ablation variants that lack one or more loss function terms, both when considering using the ground truth object category or not. The results indicate that incorporating constraints for language description parsing in the learning-based model yields more potent instructions, resulting in a more coherent attention-transferring process.

\begin{table}[t!]
    \centering
    \caption{\textbf{Grounding accuracy of \model w/ and w/o auxiliary losses on Sr3D, excluding the ``between'' relationship.}}
    \vspace{-6pt}
    \label{tab:loss_ablation}
    \resizebox{0.62\linewidth}{!}{%
        \begin{tabular}{ccc}
        \toprule
        \multicolumn{2}{c}{Method} & Sr3D \\
        \hline
        \multirow{5}{*}{w/o GT} & w/o $L_t$ & $59.7\%$ \\
                                & w/o $L_a$ & $54.2\%$ \\
                                & w/o $L_r$ & $57.1\%$ \\
                                & w/o aux. loss & $36.3\%$ \\
                                & w/  aux. loss & $\mathbf{62.2\%}$ \\
        \hline
        \multirow{2}{*}{w/ GT}  & w/o aux. loss & $96.4\%$ \\
                                & w/  aux. loss & $\mathbf{99.1\%}$ \\
        \bottomrule
        \end{tabular}%
    }%
    \vspace{-6pt}
\end{table}

\section{Conclusion} \label{sec:conclusion}
In this work, we propose \model, a novel neural-symbolic model, to address the \acl{3dvg} problem in a reasoning manner.
\model represents the 3D scene into a scene graph and parses the language description into instruction, using a shared semantic concept vocabulary.
We formulate the grounding process as a human-like reasoning process on the scene graph rather than a score-matching mechanism, improving the model's interpretability and generalization ability. Our work pioneers a new paradigm for \ac{3dvg}, potentially igniting further innovation and inspiration. 

\paragraph{Limitations and future works}
Since the scene graph is intractable to modeling the spatial relationships among multiple objects, we neglect the ternary relationship ``between'', which is frequently encountered in referential utterances. Besides, as demonstrated in \cref{sec:ablation}, the grounding accuracy is impeded by the low object classification accuracy. In the future, we will direct our efforts toward addressing these two issues.

\clearpage
{
\bibliographystyle{named}
\bibliography{ijcai24}
}

\end{document}